\documentclass[10pt,twocolumn,letterpaper]{article}

\usepackage{cvpr}
\usepackage{times}
\usepackage{epsfig}
\usepackage{graphicx}
\usepackage{subcaption}
\usepackage{amsmath}
\usepackage{amssymb}

\usepackage[pagebackref=true,breaklinks=true,letterpaper=true,colorlinks,bookmarks=false]{hyperref}

\cvprfinalcopy % *** Uncomment this line for the final submission

\def\cvprPaperID{6108} % *** Enter the CVPR Paper ID here

\ifcvprfinal\fi
\begin{document}

%%%%%%%%% TITLE
\title{Automatic Dataset Annotation to Learn CNN Pore \\ Description for Fingerprint Recognition}

% Authors at the same institution
\author{Gabriel Dahia \hspace{2cm} Maur\' icio Pamplona Segundo \\
Department of Computer Science, Federal University of Bahia \\
{\tt\small gdahia@gmail.com, mauriciops@ufba.br} \\
}

\maketitle
\ifcvprfinal\thispagestyle{empty}\fi

%%%%%%%%% ABSTRACT
\begin{abstract}
  High-resolution fingerprint recognition often relies on sophisticated matching algorithms based on hand-crafted keypoint descriptors, with pores being the most common keypoint choice.
  Our method is the opposite of the prevalent approach: we use instead a simple matching algorithm based on robust local pore descriptors that are learned from the data using a CNN.
  In order to train this CNN in a fully supervised manner, we describe how the automatic alignment of fingerprint images can be used to obtain the required training annotations, which are otherwise missing in all publicly available datasets.
  This improves the state-of-the-art recognition results for both partial and full fingerprints in a public benchmark.
  To confirm that the observed improvement is due to the adoption of learned descriptors, we conduct an ablation study using the most successful pore descriptors previously used in the literature.
  All our code is available 
  \ifcvprfinal
  at \url{https://github.com/gdahia/high-res-fingerprint-recognition}.
  \else
  in the Supplementary Materials.
  \fi
\end{abstract}

%%%%%%%%% BODY TEXT
\section{Introduction}

Automated fingerprint recognition systems mostly work by detecting keypoints, usually minutiae or pores, and matching them across images.
Sometimes, the matching approach uses local descriptors to describe the keypoints' neighborhood and establish correspondences~\cite{jain-latent,direct-pore,ridge-reconstruction}.
These approaches usually provide excellent recognition results and can mostly be categorized into two settings.
The first one uses low-resolution sensors but requires high-quality images comprising the entire fingerprint~\cite{jain-pores-and-ridges,ridge-reconstruction}.
The second setting requires high-resolution images and relies on sophisticated and computationally expensive algorithms to match keypoint descriptors~\cite{ridge-reconstruction,su-pores-deep,td-sparse}.
The latter usually provides better recognition results, thus allowing the use of partial fingerprints while maintaining the system secure.
The focus of this work is on the second scenario, high-resolution fingerprint recognition.
Specifically, we consider pores as keypoints, as many works in the literature do~\cite{direct-pore, ridge-reconstruction, td-sparse, feature-guided, su-pores-deep}.

We attribute the previously mentioned need for sophisticated matching algorithms to the widespread use of hand-crafted keypoint descriptors.
These descriptors do not require training data and are fast to compute~\cite{direct-pore,ridge-reconstruction}, but are not very robust to some of the variations to which fingerprint images are susceptible, \eg the opening and closing of pores and distortions caused by the finger pressure applied to the sensor surface.
This causes many spurious keypoint correspondences when using simple matching criteria.

There were two ways to solve this problem.
The first one involves improving descriptors.
While there are approaches that learn keypoint descriptors with better results than hand-crafted ones~\cite{l2net,hardnet,doap}, their use is incipient in the fingerprint recognition literature~\cite{jain-latent, zhang-pattern-rec}.
This is probably caused by the lack of adequate keypoint annotations in publicly available fingerprint datasets, a requirement for the known approaches to learning local descriptors.
We believe this led researchers to focus on developing more and more sophisticated matching algorithms, the second way to address the spurious correspondences problem~\cite{td-sparse, ridge-reconstruction, feature-guided}.
A detailed analysis of previous works is given in Section~\ref{sec:related-work}.

In this work, we approach this problem using the former approach.
First, we show that by aligning training images, we can automatically generate the required keypoint annotations for them.
This allows learning local descriptors for pores with a convolutional neural network (CNN) in a fully supervised manner.
While it is common to find previous works that align fingerprint images, they do so as a step in the matching stage to discard spurious correspondences~\cite{direct-pore, td-sparse}.
Our work, on the other hand, uses fingerprint alignment to generate training data.
As we do this offline, and not as part of the matching method, it does not affect the computational cost to perform recognition.
Second, we show that using the proposed descriptors with a simple matching algorithm improves the state-of-the-art performance in the standard high-resolution fingerprint recognition benchmark: the Hong Kong Polytechnic University High-Resolution-Fingerprint (PolyU-HRF) dataset~\cite{direct-pore}.

Our method for fingerprint recognition consists of describing each pore with the learned descriptors, finding pore correspondences across images, and determining if the number of these correspondences is above a fixed threshold.

The contributions of this work are:
\begin{itemize}
\item we detail how aligning fingerprint images in a training set provides the required annotations to train a CNN that learns local keypoint descriptors (Section~\ref{sec:pore-description});

\item we describe our fingerprint recognition algorithm, which combines a simple matching criterion with the learned descriptors (Section~\ref{sec:rec});
%"we describe our fingerprint recognition algorithm, which combines a simple matching criterion with the descriptors learned from data annotated using our method (Section~\ref{sec:rec});" or something like it, to emphasize here in the contributions, that we use the method in our paper. maybe in the previous item or not at all?

\item we show that the proposed method yields state-of-the-art performance in the standard benchmark for the task (Section~\ref{sec:exps}).
  To verify our claim that the need for sophisticated matching is due to the use of hand-crafted descriptors, we conduct an ablation study with the most successful pore descriptors used in previous works.
\end{itemize}

%---------------------------------------------------------------------
\section{Related work}
\label{sec:related-work}

%---------------------------------------------------------------------
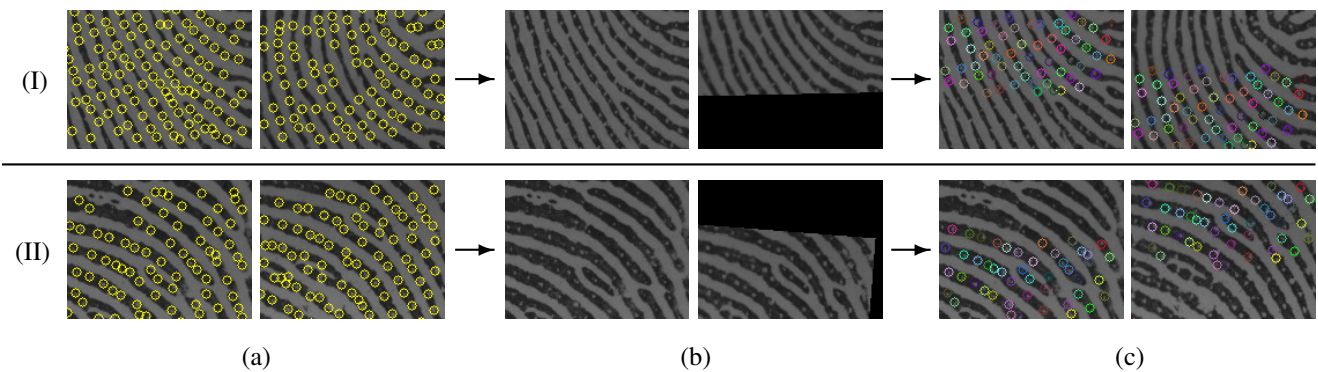
\begin{figure*}[t]
  \begin{center}
    \setlength{\unitlength}{\textwidth}
    \begin{picture}(1, 0.2)
      \thicklines
      \put(0.015, 0.155){(I)}
      \put(0.01, 0.025){(II)}
      \put(0.05, 0.11){\includegraphics[width=0.14\textwidth]{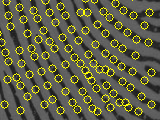}}
      \put(0.197, 0.11){\includegraphics[width=0.14\textwidth]{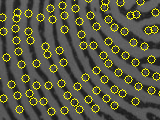}}
      \put(0.05, -0.02){\includegraphics[width=0.14\textwidth]{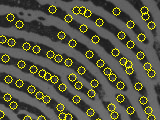}}
      \put(0.197, -0.02){\includegraphics[width=0.14\textwidth]{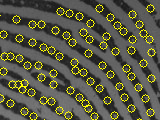}}
      \put(0.1825, -0.055){(a)}
      
      \put(0, 0.0975){\line(1, 0){1}}

      \put(0.345, 0.0325){\vector(1, 0){0.03}}
      \put(0.345, 0.1625){\vector(1, 0){0.03}}

      \put(0.383, 0.11){\includegraphics[width=0.14\textwidth]{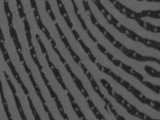}}
      \put(0.53, 0.11){\includegraphics[width=0.14\textwidth]{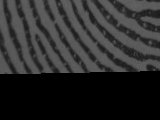}}
      \put(0.383, -0.02){\includegraphics[width=0.14\textwidth]{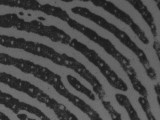}}
      \put(0.53, -0.02){\includegraphics[width=0.14\textwidth]{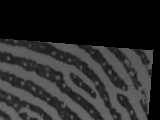}}
      \put(0.5165, -0.055){(b)}

      \put(0.677, 0.0325){\vector(1, 0){0.03}}
      \put(0.677, 0.1625){\vector(1, 0){0.03}}

      \put(0.714, 0.11){\includegraphics[width=0.14\textwidth]{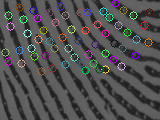}}
      \put(0.86, 0.11){\includegraphics[width=0.14\textwidth]{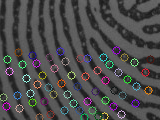}}
      \put(0.714, -0.02){\includegraphics[width=0.14\textwidth]{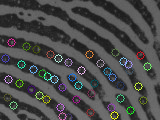}}
      \put(0.86, -0.02){\includegraphics[width=0.14\textwidth]{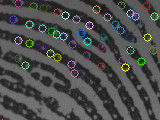}}
      \put(0.8475, -0.055){(c)}

    \end{picture}
  \end{center}
  \vspace{14pt}
  \caption{Automatic pore annotation pipeline.
  (a) A fingerprint dataset for which images have subject labels is the input to our method.
  Images in the same row belong to the same subject, \ie images in the first row belong to subject (I), images in the second row belong to subject (II).
  (b) For each subject, a reference image is picked arbitrarily and every other image of that subject is aligned to it.
  The left images are the references for their subjects.
  The right images are transformed only to show the found alignment.
  (c) Pores that are not visible in some image of its subject are discarded. The remaining pores are annotated with unique labels.
  In these images, circles of different colors correspond to different, unique pore labels.
  Though there are colors in one image that look the same, they represent different labels.
  Notice how the alignment is used only to find pore correspondences; the images are unchanged by the end of the annotation process.
  Best viewed in color.
  }
  \label{fig:annotation}
\end{figure*}
%---------------------------------------------------------------------

We focus our review on fingerprint recognition works that use pores as keypoints, PolyU-HRF as benchmark, or that learn keypoint descriptors.

Direct Pore descriptors (hereon called DP descriptors for simplicity) are obtained by normalizing fingerprint patches around pores for contrast and rotation variations~\cite{direct-pore}.
This means that using such descriptors requires computing the fingerprint orientation at every pore location.
In the first work that uses them, first, mutual nearest neighbors are used as correspondences.
This initial correspondence set is filtered using Random Sample Consensus (RANSAC)~\cite{ransac} and the final number of correspondences is used as score.

The matching algorithm of this method was improved by using tangent distance and sparse representations to find the initial set of correspondences and replacing RANSAC with weighted RANSAC (WRANSAC)~\cite{wransac}.
The overall method is called TDSWR~\cite{td-sparse}.
While it greatly improves recognition results, the use of sparse representations in the correspondence establishment step requires solving an NP-Hard optimization problem~\cite{sparse-is-hard}.
The authors approach it by solving instead an $\ell_1$-regularized least squares problem.
This surrogate problem can be solved polynomially, but it forfeits theoretical guarantees of sparse representations and still involves solving a linear program with many variables and constraints.

The latest improvement of DP-based approaches is the use of pores in the neighborhood of minutiae.
This method consists of hierarchically combining the matching scores of TDSWR obtained using two different subsets of pores.
The first set consists of all pores; the second one only considers pores that are spatially close to minutiae, which the authors call distinctive pores~\cite{feature-guided}.
This approach indirectly uses minutiae to recognize fingerprints, as local descriptors of pores that are close to minutiae can be seen as descriptors of the minutiae themselves.
Other methods also report better scores when combining pore-only matching scores with that of minutiae~\cite{direct-pore, td-sparse}.

Affine Fourier Moment-Matching (AFMM)~\cite{su-pores-deep} considers solving a different optimization problem to match fingerprints.
It aims to minimize the energy cost function between the second-degree Fourier moments of the input fingerprint images using gradient descent.
The first variant of this method, AFMM-G, considers the sums of the energy cost functions for ridge patterns, minutiae, and pores.
The second and better variant is called AFMM-B.
It proposes to first divide the gallery fingerprint image into non-overlapping blocks.
Then, for each block, a sliding window iterates over the query fingerprint to find the image region with the smallest energy cost function.
If this match is below a certain threshold, a correspondence is established.
The matching score is the total number of correspondences.
AFMM-B, then, requires that for every block and every possible sliding window position, an optimization problem be solved with gradient descent.
For partial fingerprints, the authors use 12 blocks.
They do not report results for full fingerprint images.

Another approach altogether involves describing pores using SIFT descriptors~\cite{sift} to establish pore correspondences.
This method discards spurious correspondences by reconstructing fingerprint ridges using pore detections.
Matches are scored, then, using a spatially informed metric over pairs of correspondences~\cite{ridge-reconstruction}.

The first approach, to the best of our knowledge, to learn local keypoint descriptors for fingerprint images was used for latent fingerprints and uses minutiae as keypoints~\cite{jain-latent}.
It trains a CNN patch descriptor in a dataset of rolled fingerprint images. 
To establish minutiae correspondences for training, it uses traditional fingerprint minutiae matching methods, discarding low-confidence correspondences as spurious.
One disadvantage of this approach, when compared to aligning fingerprints, is that might allow false minutiae correspondences that have high-confidence scores to be in the training set.
We also believe that this approach might not add hard minutiae correspondences to the training set because traditional minutiae matching algorithms are sub-optimal. % replace sub-optimal here

GlobalNet~\cite{zhang-pattern-rec} is a CNN trained to generate global image descriptors for 2000dpi partial fingerprint images.
Fingerprint recognition is done by determining if image descriptors for the input images are sufficiently similar.

MinutiaeNet~\cite{zhang-pattern-rec} claims to do something very similar to what we do in this work: describing local neighborhoods of keypoints and match partial fingerprints by finding correspondences amongst the generated descriptors.
What it actually does is, first, cropping ${320 \times 320}$ patches around the center of each minutia.
The patches are then described using a CNN; recognition is done using all the resulting descriptors.
We believe this setup is better framed as performing recognition using an ensemble of global image descriptors, in which each descriptor is obtained from random crops of the original image, than matching with local keypoint descriptors.
In the global image descriptor setting, almost the entirety of the image is visible in each described sub-image.
Each of the MinutiaNet patches contains more than a fourth of the entire source image, which has dimensions ${480 \times 800}$, and there are more than four minutiae per image.
For the local keypoint descriptor setting, on the other hand, sub-images are restricted to small image regions around the keypoints.
While we admit that there is no exact definition of what ``small image regions'' means in this context, we also remark that using an ensemble of global image descriptors is known to improve recognition~\cite{vggface}.

Another similarity to our work is that MinutiaNet is trained with triplet loss, albeit with modifications proposed by its authors.
The authors defend that these modifications are necessary to improve training convergence, but we did not find that that was necessary to train our CNN.

Finally, the results of the last two methods, GlobalNet and MinutiaNet, are not reproducible.
This is so because the only dataset for which they report results is not publicly available.
Besides that, we argue that the evaluation protocol we follow is more rigorous.
While the dataset used in those experiments has 1,800 images, more than PolyU-HRF, the authors split it in a way that 90\% of the images are in the training set.
This split is considerably easier than PolyU-HRF's because it not only provides a much bigger training set - 1620 against 210 - but it also tests in fewer images and subject identities - 180 against 1480.
Furthermore, the number of subjects exclusively in the test set in PolyU-HRF, at least 113, is greater than the total number of subjects in their test set, 30.

%----------------------------------------------------------------------
\section{CNN pore description}
\label{sec:pore-description}
In this work, we aim to train a CNN to describe pores such that multiple instances of a same pore have more similar descriptors than two distinct pores.
It is important to notice that subject labels do not coincide with pore labels: a single subject has many pores, and, therefore, many pore labels per image.

Since the learning approach we employ requires supervision, we must have an annotated dataset.
In this situation, this means that every pore must receive a unique label.
No fingerprint image dataset provides such annotations, to the best of our knowledge.
We believe this is because such labeling task would require workers to annotate hundreds of pores per image and a few images per subject, which would be expensive, time-consuming and burdensome.

Thus, we devised an automated procedure for annotating pores that works by aligning fingerprint images.
Figure~\ref{fig:annotation} illustrates its overall structure.
Images are iteratively aligned to a reference image that belongs to the same subject.
This alignment process uses hand-crafted descriptors to establish pore correspondences and it also enforces spatial coherence using the alignment of the previous iteration.
The final alignment allows discarding pores outside the overlap area and assigning unique labels to the remaining ones.

As we stated before, establishing correspondences based on hand-crafted descriptors causes many spurious correspondences.
Since we are specifically annotating a dataset to learn a robust descriptor, such a descriptor is not yet available to be used for alignment.
However, the previous methods are not unsuited to alignment, because it can be more costly to run and is an easier task than matching.
Alignment can be more computationally costly because it is only executed once, to annotate the training data for the CNN.
It is the trained CNN that will be used for matching.
Matching is a harder task than alignment because it is required to work for both positive image pairs, \ie images that belong to the same subject, and negative pairs, \ie images that do not belong to the same subject.
Alignment, on the other hand, is required to work only on positive pairs.
We describe the annotation process in detail next.

%----------------------------------------------------------------------
\subsection{Automatic pore annotation}
\label{sec:alignment}

Since for alignment we are using hand-crafted descriptors, preprocessing is essential.
First, we apply a median blur, to denoise the image, and we equalize the images' contrast with CLAHE~\cite{clahe}.
Both are established preprocessing steps in the fingerprint community~\cite{ridge-reconstruction}. % add another ref

% MUST MENTION THAT NOTHING IN THIS SECTION MODIFIES IMAGES IN THE TRAINING SET AND THAT THE IMAGES OF THE TEST SET NEVER TAKE PART IN THIS PROCESS.
% THIS IS ONLY FOR TRAINING DATA ANNOTATION

After, we compute SIFT~\cite{sift} descriptors for each pore.
SIFT descriptors have been previously used to perform pore-based fingerprint recognition with good results~\cite{ridge-reconstruction}.
While there are other domains in which SIFT was sometimes shown to be inferior to other techniques, it was the first descriptor we tried.
Since we were successful and aligning fingerprint images is not the purpose of this paper, we only tried SIFT.

First, we find correspondences between SIFT descriptors in images $I_1$ and $I_2$ using the original SIFT criterion: descriptors must be mutual nearest neighbors and pass a distance ratio check for their second nearest neighbors~\cite{sift}.

With this initial set of correspondences, we align $I_1$ and $I_2$ using the closed form solution to the absolute orientation problem~\cite{horn}.
The overlap between images is determined using the alignment and pores outside of it are discarded.
We repeat this process, but in iteration ${i + 1}$ we consider the following metric for pores $p_1$ of $I_1$ and $p_2$ of $I_2$ to find descriptor correspondences:
\begin{equation}
  \label{eq:alignment}
  d_{i + 1}(p_1, p_2) = d_{\mathit{SIFT}}(p_1, p_2) + \frac{\lambda}{w_i + \epsilon} ||p_1 - T_i(p_2)||^2,
\end{equation}
where $d_\mathit{SIFT}$ is the distance between SIFT descriptors, $T_i$ is the alignment transformation in iteration $i$, $w_i$ is the mean squared error (MSE) of the correspondences found in iteration $i$,  $\lambda$ is a user specified parameter, and $\epsilon$ is used for numerical stability.
This process stops in convergence or when the maximum number of iterations $t$ is reached.

The second term in Equation~\ref{eq:alignment} represents the Euclidean distance, in aligned space, between the 2D coordinates of pores $p_1$ and $p_2$.
We use this term to determine how good the current alignment is.
The intuition behind this thinking is that a perfect alignment should have zero MSE.
So, when the alignment has high MSE, the images are not yet well aligned.
This means that the similarity of pore descriptors will outweigh the Euclidean distance between the pores when establishing correspondences.
Similarly, when the alignment has low MSE, the images are better aligned.
Thus, correspondences with similar descriptors that do not fit the current alignment are unlikely to be created, reducing spurious correspondences.
This rationale is encoded in Equation~\ref{eq:alignment} by $w_i$, with $\lambda$ serving to control the proportion between $d_\mathit{SIFT}$ and Euclidean distance.

% why can alignment use hand-crafted descriptors and matching cannot? iterative with spatial information discards false correspondences, but is costly
% state that alignment need only work for positive pairs and that is costly. since it is only done once, it is ok

\begin{figure}[h]
  \begin{center}
    \begin{subfigure}{0.48\linewidth}
      \includegraphics[width=\textwidth]{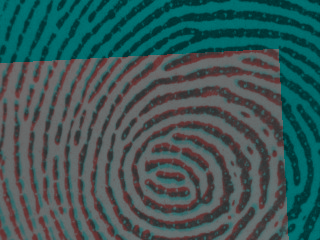}
      \caption{\label{fig:align1}}
    \end{subfigure}
    \begin{subfigure}{0.48\linewidth}
      \includegraphics[width=\textwidth]{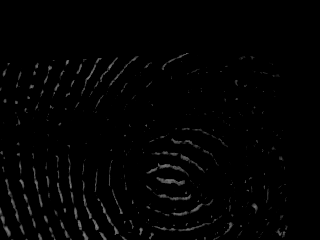}
      \caption{\label{fig:align2}}
    \end{subfigure}
  \end{center}
  \caption{Aligned fingerprint images.
  (a) Aligned images are superimposed in different color channels.
  (b) Only the alignment residues for one of the images are shown.
  These images show that fingerprints are not perfectly alignable by rigid transformations.
  Best viewed in color.
  }
  \label{fig:align}
\end{figure}

Having the images aligned does not solve the dataset annotation problem in its entirety.
One must also be able to determine what is the minimum distance between two keypoints so that they should have a different label.
This is an issue because two fingerprint images are often not perfectly alignable with rigid transformations.
Figure~\ref{fig:align} shows images of the same fingerprint that are not perfectly alignable due to non-rigid transformations.

%----------------------------------------------------------------------
%----------------------------------------------------------------------

To address this issue, we assume that the proposed alignment is capable of perfectly aligning images, \ie the final transformation $T$ maps coordinates between aligned images without distortion.
Hence, to generate keypoint identity annotations, we choose an arbitrary image for each subject as a reference image.
Each pore in reference images is then assigned a unique label.
Afterward, every other image is aligned to its corresponding reference image.
The coordinates for the labeled pores are then found in the other images belonging to the same subject using the alignment.
If a pore's coordinates are outside the image boundaries for some image, then this pore is discarded.
Finally, a patch of dimensions ${33 \times 33}$ is extracted around each of the remaining pores.
The set of pore patches and their corresponding labels form the training set.
Notice that the training images are aligned only in the sense that we find transformations to map coordinates from one of them into the other.
We use the transformations to annotate the data, not to transform the images themselves.

Allowing the direct use of the alignment transformation to generate pore labels is one of the benefits of aligning images to annotate the dataset.
This means that if the alignment is successful, even pores for which the hand-crafted descriptors were not able to find a confident correspondence are correctly labeled.
The other benefit is that this approach assures that the pore annotations are spatially coherent, \ie no label is assigned to a pore in a way that is spatially incompatible with the other labels assigned by the method.

%----------------------------------------------------------------------
\subsection{Training problem formulation}
\label{sec:desc-train}

Given image patches of dimensions ${33 \times 33}$ centered on pores with corresponding pore labels, our task is to train a CNN to learn pore descriptors such that descriptors for pores with the same label are closer in $\ell_2$ space than those with different ones.
HardNet~\cite{hardnet} showed state-of-the-art performance in the wide baseline stereo, patch verification, and instance retrieval tasks until recently~\cite{doap}.
We can frame the fingerprint recognition problem based on finding pore descriptor correspondences as multiple instance retrieval problems.
This allows training HardNet to learn pore descriptors in $\ell_2$ space.
Since its architecture takes as input patches with dimensions ${32 \times 32}$, we discard the last row and the last column of the pore patches.
Doing this enables us to use HardNet's architecture without modifications.

While we do not modify the network's architecture, we train it differently.
HardNet is supposed to be trained with a mini-batch containing a single positive pair of patches.
However, Schroff~\etal~\cite{facenet}, which proposes a very similar approach to learn face embeddings, state that the approach taken by HardNet is sub-optimal.
None of these works provide evidence that their approaches are better than the other, but Schroff~\etal state that unreported experiments show that triplet loss works better with more examples of the same identity in each batch.
With these considerations in mind, we employ Schroff~\etal's approach, which is called triplet semi-hard loss.

Also, to improve the robustness of our descriptors, we augment the dataset as we sample mini-batches from it.
This is done by randomly sampling a transformation that includes translating, rotating, and perturbing the contrast and brightness of each patch from a multivariate normal distribution with zero covariance.
Since we augment after we crop the patches from the image, we treat translations and rotations as if the patch is zero-padded.
We also linearly interpolate the image when the sampled transformation requires it.

It is important to notice that our procedure for augmentation is able to simulate patches undergoing rigid transformations.
This corresponds to a user translating and changing his/her finger's orientation with respect to the sensor from one capture session to another.
If rigid transformations were the only concern, we could forfeit pore identity annotation altogether and augmentation alone would suffice to provide same identity patches.
However, if a user presses his/her finger against the sensor with a different amount of pressure in a subsequent session, then there is no rigid transformation that can map the earlier image into the new one.
This is not to mention other sources of variation for fingerprint images, such as those caused by age or the presence of sweat or dirt in the finger. % give a reference for this?
This is why pore identity annotation is essential to our work.

%----------------------------------------------------------------------
\section{Fingerprint matching}
\label{sec:rec}
Given two fingerprint images $I_1$ and $I_2$ annotated with pore coordinates, the proposed method requires descriptors for each pore.
They are described using a CNN trained following the steps in Section~\ref{sec:pore-description}.
This results in descriptor set $D_1$, for the pores of image $I_1$, and descriptor set $D_2$, for the pores in image $I_2$.

It is important to notice that the alignment in Section~\ref{sec:alignment} is not part of our fingerprint matching method, nor is the use of SIFT descriptors.
These are used only in steps prior to the training of the CNN, which takes place once and is separate from the fingerprint matching method.

Next, we establish correspondences by finding mutual nearest neighbors between the descriptors in $D_1$ and in $D_2$.
This means that nearest neighbors for descriptors in $D_1$ are searched in $D_2$ and vice-versa.
We further filter these correspondences using a distance ratio check for the second nearest neighbors~\cite{sift}.

The recognition score, then, is the final number of established correspondences.
If they are above a given threshold, the pair is deemed genuine, \ie the images belong to the same fingerprint; otherwise, the pair is considered an impostor pair, \ie they do not belong to the same fingerprint.

Our approach to fingerprint matching is considerably simpler than previous works.
The proposed method requires aligning only training images and this only happens once, prior to the training of the CNN.
Previous methods require alignment, either explicitly or implicitly, \eg with RANSAC, for matching~\cite{direct-pore, td-sparse}.
Other methods require solving optimization problems or reconstructing fingerprint ridges~\cite{su-pores-deep, feature-guided, ridge-reconstruction}.

%----------------------------------------------------------------------
\section{Experiments and results}
\label{sec:exps}
We use the PolyU-HRF dataset~\cite{direct-pore}, the public benchmark for high-resolution fingerprint recognition, for our experiments.

We conduct experiments to (1) compare our results to previous methods, and (2) determine if found differences are due to improved descriptors.
To assess (2), we conduct an ablation study, keeping the entire matching method unchanged except for the used pore descriptors.
We replace the proposed descriptors for each descriptor previously used in the literature, comparing the observed changes.

The pore description CNN is optimized using Stochastic Gradient Descent (SGD)~\cite{sgd} with early stopping.
Our implementation uses Tensorflow~\cite{tf}, NumPy~\cite{np}, and OpenCV~\cite{opencv}.
We make all the code required to train the CNN and to reproduce our experiments, alongside the trained model, publicly available.

%----------------------------------------------------------------------
\subsection{PolyU-HRF dataset}
\label{sec:dataset}
Polyu-HRF is divided into two subsets: \textit{DBI} and \textit{DBII}.

\textit{DBI} consists of 210 images of partial fingerprints, with 6 images per subject, in a training set and 1,480 images, with 10 images per subject, in a test set.
It provides the corresponding identity for each of those images and separates them in acquisition sessions.

\textit{DBII} has no training set and its test set has the same number of images and subjects of \textit{DBI}'s.
The difference is that, in \textit{DBII}, the images are of full-size fingerprints, instead of partial ones.
Its label information is like that of \textit{DBI}.

PolyU-HRF's \textit{DBI} and \textit{DBII} do not provide pore coordinate annotations, and, therefore, lack pore identity annotations for its images.
We believe no fingerprint dataset provides this information and that this is the reason why hand-crafted features are so prevalent for this task.
This also means that we must detect pores in the images to conduct fingerprint recognition experiments.

%----------------------------------------------------------------------
\subsection{Experimental setup}

%----------------------------------------------------------------------
\begin{figure*}
    \begin{subfigure}{0.5\linewidth}
      \includegraphics[width=\textwidth]{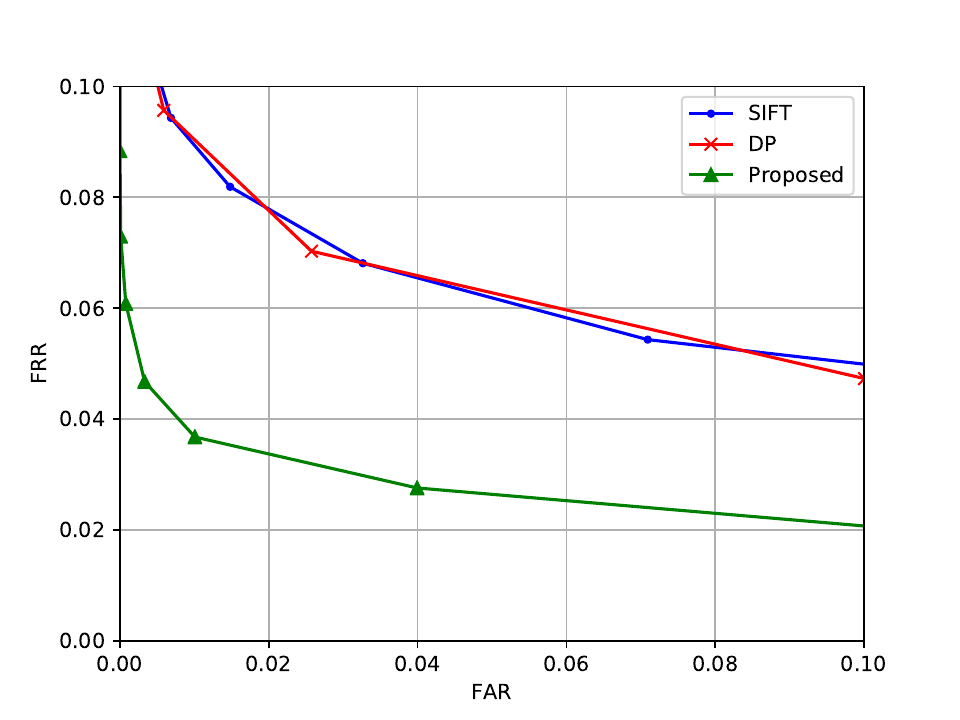}
      \caption{\label{fig:roc1}}
    \end{subfigure}
    \begin{subfigure}{0.5\linewidth}
      \includegraphics[width=\textwidth]{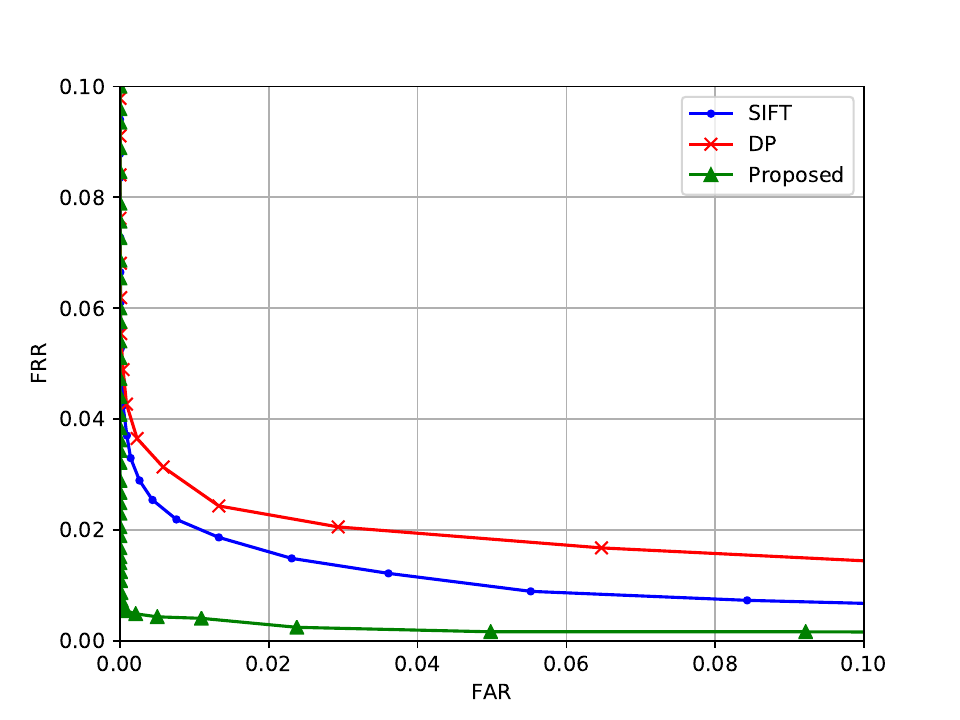}
      \caption{\label{fig:roc2}}
    \end{subfigure}
  \caption{ROC curves for descriptor ablation study in (a) \textit{DBI} and (b) \textit{DBII}.
  Keeping the matching method unchanged and changing the descriptor shows that the proposed descriptor is responsible for the improvement in recognition results.
  This is observed in the entirety of the plot.
  Best viewed in color.}
  \label{fig:rocs}
\end{figure*}
%----------------------------------------------------------------------

We conducted fingerprint recognition experiments for the proposed method in both subsets of PolyU-HRF destined for testing.
The test set of \textit{DBI} is destined to evaluate performance in the partial fingerprints scenario, while \textit{DBII} is for evaluating the method with full fingerprint images.

We followed the established protocol for validating fingerprint recognition methods in PolyU-HRF, which was proposed by its authors and is the same for both \textit{DBI} and \textit{DBII}~\cite{direct-pore}.
This protocol consists of performing all 3,700 comparisons of images with the same identity but across sessions.
These are the only genuine comparisons to be performed.
The impostor comparisons are the 21,756 comparisons of images with different identities obtained by matching, for each subject, the first image of the first session with the first image of the second session of every other subject.

In order to determine if observed differences in our results arise from the robustness of the proposed descriptor, we conduct an ablation study.
In it, we repeat our experiment, keeping the same protocol and overall fingerprint recognition method, but replace the proposed descriptors with every other local descriptor used by previous works to describe pores.
In one experiment, we use SIFT descriptors~\cite{ridge-reconstruction} and in the other we use DP descriptors~\cite{direct-pore, td-sparse, feature-guided}.

For these other descriptors, we attempt to reproduce the local description pipeline as close as we can interpret from the work that proposes to use them as pore descriptors.
To allow a fair comparison, we optimize the parameters for these experiments with other descriptors observing the recognition Equal Error Rate (EER) in the training set.
The SIFT parameter values were scale of 8, median blur kernel dimensions of ${3 \times 3}$, and CLAHE clip limit of 3.
For DP descriptors, we used ${32 \times 32}$ patches, obtained discarding the last row and column of the ${33 \times 33}$ patches centered at each pore detection.
The distance ratio check threshold is 0.7 for both of them.

PolyU-HRF's images in \textit{DBI} and \textit{DBII} are not annotated with pore coordinates.
So, to conduct our experiments, we must first detect pores in these images.
To this end, we use a pore detection CNN\ifcvprfinal~\cite{fcn-pore-det}\else\cite{supp-fcn-pore-det}\fi with its default parameters. 

To generate the training set annotations, we empirically determined the alignment parameters by observing the results in a few training images.
Their values are ${\lambda = 500}$, ${\epsilon = 10^{-5}}$ and ${t = 10}$.
SIFT related parameters were used as determined previously, except for the distance ratio check threhsold, which was of 0.8.

We randomly split \textit{DBI}'s training images, reserving 60\% of them for an effective training set, \ie images we actually use for training the description CNN.
The remaining 40\% of the training images form a validation set.
Importantly, we make this split subject-independent, \ie there are no subject identities both in the training set and in the validation set.
Another important aspect of validation is that we perform manual hyper-parameter tuning and early stopping based directly on the end-task.
That is, we observe the recognition EER using descriptors from the model in the current training stage.
The distance ratio check is fixed at 0.7 during training to avoid unforeseen variations in the training dynamics.

The best trained model obtained recognition EER of 0\% in the validation set.
Its parameters are constant learning rate of $10^{-1}$, batches of size 252, with exactly 6 patches per pore, dropout rate of 0.3, triplet semi-hard loss margin of 2.0, and no weight decay.

While we take steps to ensure that there is no identity that appears in both the training and validation sets to improve generalization, the default protocol of PolyU-HRF has subjects that appear in both its training and test sets.
This setting is sub-optimal to evaluate methods based on deep neural networks, as it is known that they are able to memorize the training set~\cite{deep-learning-generalization}.

We believe it is unlikely that dataset memorization represents a major issue in our evaluation.
We argue this based on three facts.
First, while there are subjects in both training and test sets, there is no image in the test set that is also in the training set.
Second, our method obtains 0\% recognition EER in a validation set that has no subject overlap with the training set.
Third, despite this overlap, most subjects in the test set are not available for training.

%----------------------------------------------------------------------
\subsection{Results}

Figure~\ref{fig:rocs} presents Receiver Operating Characteristic (ROC) curves for our ablation study in \textit{DBI} and \textit{DBII}.
Table~\ref{table:ablation} displays EERs for these experiments.

%---------------------------------------------------------------------
\begin{table}[h]
  \begin{center}
    \begin{tabular}{l|c|c}
      \textbf{Descriptor}                                   & \textbf{DBI}    & \textbf{DBII} \\ \hline
      \textbf{Proposed}                                     & \textbf{3.05\%} & \textbf{0.44\%} \\ \hline
      SIFT                                                  & 5.87\%          & 1.71\% \\ \hline
      DP                                                    & 5.98\%          & 2.22\% \\
    \end{tabular}
  \end{center}
  \vspace{-4pt}
  \caption{EERs for \textit{DBI} and \textit{DBII} for the descriptor ablation study.
  Using the proposed descriptors while keeping the matching method otherwise unchanged reduces EER to almost half of the best hand-crafted descriptors' in \textit{DBI} and to almost a quarter in \textit{DBII}.
  Best results are in boldface.
  }
  \label{table:ablation}
\end{table}
%---------------------------------------------------------------------

Our ablation study provides strong evidence that the effectiveness of our approach is due to the improved descriptors.
Replacing the proposed descriptors with SIFT ones results in 5.87\% EER in \textit{DBI}; with DP ones, it results in 5.98\%.
This shows that, contrary to what was previously reported~\cite{direct-pore}, at least for this simple matching algorithm and partial fingerprints, the performance for SIFT and DP descriptors are very similar.
In contrast, when using the proposed descriptors, the EER value is only 3.05\%, cutting the error almost in half.
The effect of using learned descriptors is even more significant in \textit{DBII}: the EER, 0.44\%, is almost a quarter of SIFT's, 1.71\%, and less than a fifth of DP's, which is 2.22\%.
The ROC curves also show that for the entire plot, our results are better.

Table~\ref{table:results} compare our results with previously reported methods in terms of EER for both subsets.
We note that Su~\etal~\cite{su-pores-deep} do not report results for \textit{DBII} and that is why its entry is blank in Table~\ref{table:results}.
Our results are only worse than methods that leverage the use of minutiae when comparing EER for both the partial fingerprints in \textit{DBI} and the full fingerprints in \textit{DBII}.
This confirms that the proposed method is the state-of-the-art in pore-based fingerprint recognition for high-resolution images, according to the PolyU-HRF benchmark.

%---------------------------------------------------------------------
\begin{table}[h]
  \begin{center}
    \begin{tabular}{l|c|c}
      \textbf{Method}                                                        & \textbf{DBI}    & \textbf{DBII} \\ \hline
      Liu \etal \cite{feature-guided}*                                       & \textbf{2.17\%} & \textbf{0.17\%} \\ \hline
      \textbf{Proposed}                                                      & \textbf{3.05\%} & \textbf{0.44\%} \\ \hline
      Liu \etal \cite{td-sparse}                                             & 3.25\%          & 0.53\% \\ \hline
      Su \etal \cite{su-pores-deep}*                                         & 3.66\%          & - \\ \hline
      \ifcvprfinal Pamplona \fi Segundo \& Lemes \cite{ridge-reconstruction} & 3.74\%          & 0.76\% \\ \hline
      Zhao \etal \cite{direct-pore}*                                         & 12.40\%         & 5.93\% \\ \hline
      Zhao \etal \cite{direct-pore}                                          & 20.49\%         & 7.05\% \\
    \end{tabular}
  \end{center}
  \vspace{-4pt}
  \caption{EERs for fingerprint recognition in \textit{DBI} and \textit{DBII}.
  * marks methods that consider both minutiae and pores as keypoints.
  Our method outperforms all previous methods that use only on pores for recognition.
  We note that Su~\etal~\cite{su-pores-deep} does not report results for \textit{DBII} and that is why its entry is blank.
  Best results for methods using only pores and overall methods are in boldface.
  }
  \label{table:results}
\end{table}
%---------------------------------------------------------------------

Observe that combining minutiae information with that of pores improves the result of fingerprint recognition.
In \textit{DBI}, Zhao~\etal's~\cite{direct-pore} results go from 20.49\% to 12.40\% and Liu~\etal's~\cite{feature-guided} goes from 3.25\% to 2.17\%.
The same effect occurs in \textit{DBII}.
This is evidence that combining the proposed descriptors somehow with minutiae could yield even better results.

This comparison also confirms our hypothesis that more robust descriptors do not need the sophisticated matching methods previously used in the literature.
The previous methods either use the way in which we establish correspondences as an intermediate step~\cite{direct-pore, td-sparse, feature-guided}, involve solving complex optimization problems~\cite{td-sparse, feature-guided, su-pores-deep}, or require reconstructing fingerprint ridges~\cite{ridge-reconstruction}.
We believe this shows that they are more complex than the algorithm presented in Section~\ref{sec:rec}.

It is also interesting to remark that the only method to present better results quantified by EER than ours does so when combining already sophisticated matching methods with minutiae information.
Other methods present higher EER, even when using minutiae information for matching.

%----------------------------------------------------------------------
\section{Conclusion}

In this work, we detailed how aligning fingerprint images allows us to automatically annotate a training set with keypoint labels.
Our method iteratively establishes correspondences based on the similarity of hand-crafted keypoint descriptors and aligns positive pairs of images with them.
This process is more computationally expensive than our matching algorithm.
However, it has no impact in the running time of recognition since it only takes place once, prior to training the CNN.
Finally, to show the usefulness of that approach, we annotated the training set of the default benchmark for high-resolution fingerprint recognition with pore labels and used those labels for training the CNN we used in our experiments.

We see as our main contribution the substantial empirical evidence to corroborate our hypothesis that improving local descriptors allows the use of simple matching rules to improve high-resolution fingerprint recognition.
We both (1) obtain the state-of-the-art performance in the default benchmark for high-resolution, pore-based fingerprint recognition using a method that fulfills these criteria, (2) and conduct an ablation study with the most successful descriptors used in the literature to confirm that the improvement is due to better descriptors.
% put "210 (minus validation) images for training only!" somewhere here in the conclusion and in other places, probably the intro and when comparing to minutianet, which has 0.9*1800

% "our work could reorient the fields efforts towards improving descriptos, a largely ignored aspect of fingerprint recognition, but one we prove is essential, still unexplored and with large improvement margin"
% matching algorithms are getting close to complexity theory barriers, such as NP-Hardness. improving descriptors is a path to continuing improving fingerprint recognition without addressing such problems

Our method, expectedly, has limitations.
The most significant, in our opinion, has to do with using the data to learn the descriptor.
It is known that deep learning approaches suffer from domain variability~\cite{domain-gan}.
Fingerprint images obtained from different sensors vary considerably.
This can be seen by comparing images in PolyU-HRF~\cite{direct-pore} and in the proprietary dataset used by MinutiaNet~\cite{zhang-pattern-rec}.
The question, then, is if this could impact the performance of using a CNN trained with images from one sensor in images obtained from a different sensor.

Future work should investigate if domain variability represents a limitation of our method.
This would involve testing the trained CNN in a dataset of images from a different sensor.
As of the writing of this work, this would require collecting a new dataset or obtaining approval of use from the maintainers of a proprietary high-resolution fingerprint dataset.
PolyU-HRF is the only publicly available dataset for high-resolution fingerprints, to the best of our knowledge.
If such an investigation finds that domain variability impacts the method significantly, one could solve it in either a supervised manner, which would involve labeling fingerprints from several sensors, or an unsupervised manner, \eg using domain adversarial training~\cite{domain-gan}.

A straightforward way to improve our method would be to replace the triplet-based loss we use to train our CNN with histogram loss~\cite{histogram-loss}.
Descriptors that use this loss~\cite{doap} have recently surpassed HardNet's performance in instance retrieval and are the new state-of-the-art for this task.

Using minutiae somehow in the proposed fingerprint matching method also appears likely to be able to improve our results.
Previous work~\cite{feature-guided, td-sparse} show that this holds potential to make recognition significantly more discriminative.
This would probably involve adding steps to the matching algorithm to detect and find correspondences for minutiae but the performance improvement could outweigh the increased computational cost for high-security environments.

Another promising line of work is the use of a global fingerprint texture descriptor, like GlobalNet~\cite{zhang-pattern-rec}, either combined with or side-stepping keypoint-based recognition altogether.
This may prove difficult for the proposed PolyU-HRF protocol since it only allows 210 images to be used for training.
The original dataset does not reserve images for validation, which must then be taken from the training set.

Finally, we made the code required to reproduce our experiments publicly available.
We believe this substantially improves the transparency, reproducibility, and the usefulness of our work.

\ifcvprfinal
\section*{Acknowledgments}
This work was funded by Universidade Federal da Bahia (UFBA) and Akyiama Solu\c c\~oes Tecnol\'ogicas.
The Titan Xp we used was donated by the NVIDIA Corporation.
\fi

{\small
\bibliographystyle{ieee}
\bibliography{main}

\begin{thebibliography}{10}\itemsep=-1pt

\bibitem{tf}
M.~Abadi, P.~Barham, J.~Chen, Z.~Chen, A.~Davis, J.~Dean, M.~Devin,
  S.~Ghemawat, G.~Irving, M.~Isard, et~al.
\newblock Tensorflow: a system for large-scale machine learning.
\newblock In {\em OSDI}, volume~16, pages 265--283, 2016.

\bibitem{opencv}
G.~Bradski.
\newblock {The OpenCV Library}.
\newblock {\em Dr. Dobb's Journal of Software Tools}, 2000.

\bibitem{jain-latent}
K.~Cao and A.~K. Jain.
\newblock Automated latent fingerprint recognition.
\newblock {\em IEEE Transactions on Pattern Analysis and Machine Intelligence},
  2018.

\bibitem{fcn-pore-det}
G.~Dahia and M.~P. Segundo.
\newblock Improving fingerprint pore detection with a small fcn.
\newblock {\em arXiv preprint arXiv:1811.06846}, 2018.

\bibitem{sparse-is-hard}
G.~Davis, S.~Mallat, and M.~Avellaneda.
\newblock Adaptive greedy approximations.
\newblock {\em Constructive Approximation}, 13(1):57--98, Mar 1997.

\bibitem{ransac}
M.~A. Fischler and R.~C. Bolles.
\newblock Random sample consensus: a paradigm for model fitting with
  applications to image analysis and automated cartography.
\newblock {\em Communications of the ACM}, 24(6):381--395, 1981.

\bibitem{domain-gan}
Y.~Ganin, E.~Ustinova, H.~Ajakan, P.~Germain, H.~Larochelle, F.~Laviolette,
  M.~Marchand, and V.~Lempitsky.
\newblock Domain-adversarial training of neural networks.
\newblock {\em The Journal of Machine Learning Research}, 17(1):2096--2030,
  2016.

\bibitem{doap}
K.~He, Y.~Lu, and S.~Sclaroff.
\newblock Local descriptors optimized for average precision.
\newblock In {\em The IEEE Conference on Computer Vision and Pattern
  Recognition (CVPR)}, June 2018.

\bibitem{horn}
B.~K. Horn, H.~M. Hilden, and S.~Negahdaripour.
\newblock Closed-form solution of absolute orientation using orthonormal
  matrices.
\newblock {\em JOSA A}, 5(7):1127--1135, 1988.

\bibitem{jain-pores-and-ridges}
A.~K. Jain, Y.~Chen, and M.~Demirkus.
\newblock Pores and ridges: High-resolution fingerprint matching using level 3
  features.
\newblock {\em IEEE Transactions on Pattern Analysis and Machine Intelligence},
  29(1):15--27, Jan 2007.

\bibitem{td-sparse}
F.~Liu, Q.~Zhao, and D.~Zhang.
\newblock A novel hierarchical fingerprint matching approach.
\newblock {\em Pattern Recognition}, 44(8):1604--1613, 2011.

\bibitem{feature-guided}
F.~Liu, Y.~Zhao, and L.~Shen.
\newblock Feature guided fingerprint pore matching.
\newblock In {\em Chinese Conference on Biometric Recognition}, pages 334--343.
  Springer, 2017.

\bibitem{sift}
D.~G. Lowe.
\newblock Object recognition from local scale-invariant features.
\newblock In {\em Computer vision, 1999. The proceedings of the seventh IEEE
  international conference on}, volume~2, pages 1150--1157. Ieee, 1999.

\bibitem{hardnet}
A.~Mishchuk, D.~Mishkin, F.~Radenovic, and J.~Matas.
\newblock Working hard to know your neighbor's margins: Local descriptor
  learning loss.
\newblock In {\em Advances in Neural Information Processing Systems}, pages
  4826--4837, 2017.

\bibitem{np}
T.~E. Oliphant.
\newblock {\em A guide to NumPy}, volume~1.
\newblock Trelgol Publishing USA, 2006.

\bibitem{ridge-reconstruction}
M.~Pamplona~Segundo and R.~de~Paula~Lemes.
\newblock Pore-based ridge reconstruction for fingerprint recognition.
\newblock In {\em The IEEE Conference on Computer Vision and Pattern
  Recognition (CVPR) Workshops}, June 2015.

\bibitem{vggface}
O.~M. Parkhi, A.~Vedaldi, and A.~Zisserman.
\newblock Deep face recognition.
\newblock In {\em British Machine Vision Conference}, 2015.

\bibitem{sgd}
H.~Robbins and S.~Monro.
\newblock A stochastic approximation method.
\newblock In {\em Herbert Robbins Selected Papers}, pages 102--109. Springer,
  1985.

\bibitem{facenet}
F.~Schroff, D.~Kalenichenko, and J.~Philbin.
\newblock Facenet: A unified embedding for face recognition and clustering.
\newblock In {\em Proceedings of the IEEE conference on computer vision and
  pattern recognition}, pages 815--823, 2015.

\bibitem{su-pores-deep}
H.-R. Su, K.-Y. Chen, W.~J. Wong, and S.-H. Lai.
\newblock A deep learning approach towards pore extraction for high-resolution
  fingerprint recognition.
\newblock In {\em Acoustics, Speech and Signal Processing (ICASSP), 2017 IEEE
  International Conference on}, pages 2057--2061. IEEE, 2017.

\bibitem{l2net}
Y.~Tian, B.~Fan, and F.~Wu.
\newblock L2-net: Deep learning of discriminative patch descriptor in euclidean
  space.
\newblock In {\em 2017 IEEE Conference on Computer Vision and Pattern
  Recognition (CVPR)}, pages 6128--6136, July 2017.

\bibitem{histogram-loss}
E.~Ustinova and V.~Lempitsky.
\newblock Learning deep embeddings with histogram loss.
\newblock In {\em Advances in Neural Information Processing Systems}, pages
  4170--4178, 2016.

\bibitem{deep-learning-generalization}
C.~Zhang, S.~Bengio, M.~Hardt, B.~Recht, and O.~Vinyals.
\newblock Understanding deep learning requires rethinking generalization.
\newblock {\em arXiv preprint arXiv:1611.03530}, 2016.

\bibitem{wransac}
D.~Zhang, W.~Wang, Q.~Huang, S.~Jiang, and W.~Gao.
\newblock Matching images more efficiently with local descriptors.
\newblock In {\em Pattern Recognition, 2008. ICPR 2008. 19th International
  Conference on}, pages 1--4. Citeseer, 2008.

\bibitem{zhang-pattern-rec}
F.~Zhang, S.~Xin, and J.~Feng.
\newblock Combining global and minutia deep features for partial
  high-resolution fingerprint matching.
\newblock {\em Pattern Recognition Letters}, 2017.

\bibitem{direct-pore}
Q.~Zhao, L.~Zhang, D.~Zhang, and N.~Luo.
\newblock Direct pore matching for fingerprint recognition.
\newblock In {\em International Conference on Biometrics}, pages 597--606.
  Springer, 2009.

\bibitem{clahe}
K.~Zuiderveld.
\newblock Contrast limited adaptive histogram equalization.
\newblock {\em Graphics gems}, pages 474--485, 1994.

\end{thebibliography}
}

\end{document}